\documentclass{svproc}
\usepackage{url}
\usepackage{tikz}
\usepackage{graphicx}
\usepackage{mathtools}
\usepackage{amsmath}
\usepackage{amssymb}

\usepackage{algorithm}
\usepackage{algorithmic}
\usepackage{hyperref}
\begin{document}
\mainmatter              
\title{Safe Reinforcement Learning-Based Vibration Control: Overcoming Training Risks with LQR Guidance}
\titlerunning{10th International Congress on
Computational Mechanics and Simulation}  
%
\author{Rohan Vitthal Thorat\inst{1}, Juhi Singh\inst{2} \and
Rajdip Nayek\inst{1}}
\authorrunning{R.V. Thorat et al.} 
%
%
\institute{Applied Mechanics department, IIT Delhi, India\\
\and
Mechanical Engineering department, IIT Delhi, India}

\maketitle              

\begin{abstract}
Structural vibrations induced by external excitations pose significant risks, including safety hazards for occupants, structural damage, and increased maintenance costs. While conventional model-based control strategies like Linear Quadratic Regulator (LQR) and Model Predictive Control (MPC) effectively mitigate vibrations, their reliance on accurate system models necessitates tedious system identification. This tedious system identification process can be avoided by using model-free Reinforcement learning (RL) method for vibration control task. 
 
RL controllers derive their policies solely from observed structural behaviour, eliminating the requirement for an explicit structural model. For an RL controller to be truly model-free, its training must occur on the actual physical system rather than in simulation. However, during this training phase, the RL controller lacks prior knowledge and it exerts control force on the structure randomly, which 
can potentially harm the structure. To mitigate this risk, we propose guiding the RL controller using a Linear Quadratic Regulator (LQR) controller. While LQR control typically relies on an accurate structural model for optimal performance, our observations indicate that even an LQR controller based on an entirely incorrect model outperforms the uncontrolled scenario. Motivated by this finding, we 
introduce a hybrid control framework that integrates both LQR and RL controllers. In this approach, the LQR policy is derived from a randomly selected model and its parameters. As this LQR policy does not require knowledge of the true or an approximate structural model the overall framework remains model-free. This hybrid approach eliminates dependency on explicit system models while minimizing 
exploration risks inherent in naive RL implementations. As per our knowledge, this is the first study to address the critical training safety challenge of RL-based vibration control and provide a validated solution.
\end{abstract}
\section{Introduction}
In last few decades, the construction of large-scale structures such as high-rise buildings and long-span bridges has rapidly expanded worldwide. With the growth of modern infrastructure, ensuring the reliability and safety of these structures has become increasingly critical. Structural reliability plays a key role not only in reducing economic losses but also in safeguarding public safety and enhancing quality of life. One of the major challenges in maintaining structural reliability is mitigating unwanted vibrations, which may arise from dynamic loads such as wind gusts, earthquakes, or human activity.  

Vibration control strategies are typically classified into \textit{passive} and \textit{active} methods. Passive control techniques rely on structural modifications or devices such as tuned mass dampers, base isolators, or viscoelastic materials to reduce vibrations, without requiring external energy input. While effective in many cases, passive systems are difficult to retrofit into existing structures since they often demand changes to the architectural or structural design. Given that the number of existing structures far exceeds the number of new constructions each year, developing control techniques suitable for retrofitting is of paramount importance.  

In contrast, active control methods apply external forces through actuators using feedback from sensors. This approach does not require fundamental changes to the underlying structure, making it more feasible for existing systems. Over the years, state-of-the-art active control strategies such as Linear Quadratic Regulator (LQR), Linear Quadratic Gaussian (LQG), and Model Predictive Control (MPC) have been widely studied. However, their effectiveness strongly depends on the accurate identification of the system’s mathematical model. Model identification requires expert knowledge, extensive instrumentation, and detailed structural data, which not only increases practical costs but also introduces uncertainties. Moreover, while MPC offers significant advantages in performance, its high computational burden often hinders real-time implementation for vibration-sensitive applications.  

These limitations motivate the need for \textit{model-free and computationally efficient control strategies} for structural vibration mitigation. Reinforcement Learning (RL) has emerged as a promising alternative, offering the ability to learn control policies directly from data without requiring explicit system models. Recent studies have applied RL to vibration control problems, demonstrating its potential for multi-degree-of-freedom systems and hybrid algorithm designs. For instance, Eshkevari et al. \cite{eshkevari2023active} investigated RL-based active vibration control, while Panda et al. \cite{panda2024continuous} proposed a hybrid algorithm combining Deep Deterministic Policy Gradient (DDPG) with Iterative Gradient-Based State Feedback Control (SFSC) \cite{panda2024iterative}, achieving improved controllability and stability.  

\begin{figure}
    \centering
    \includegraphics[width=1\linewidth]{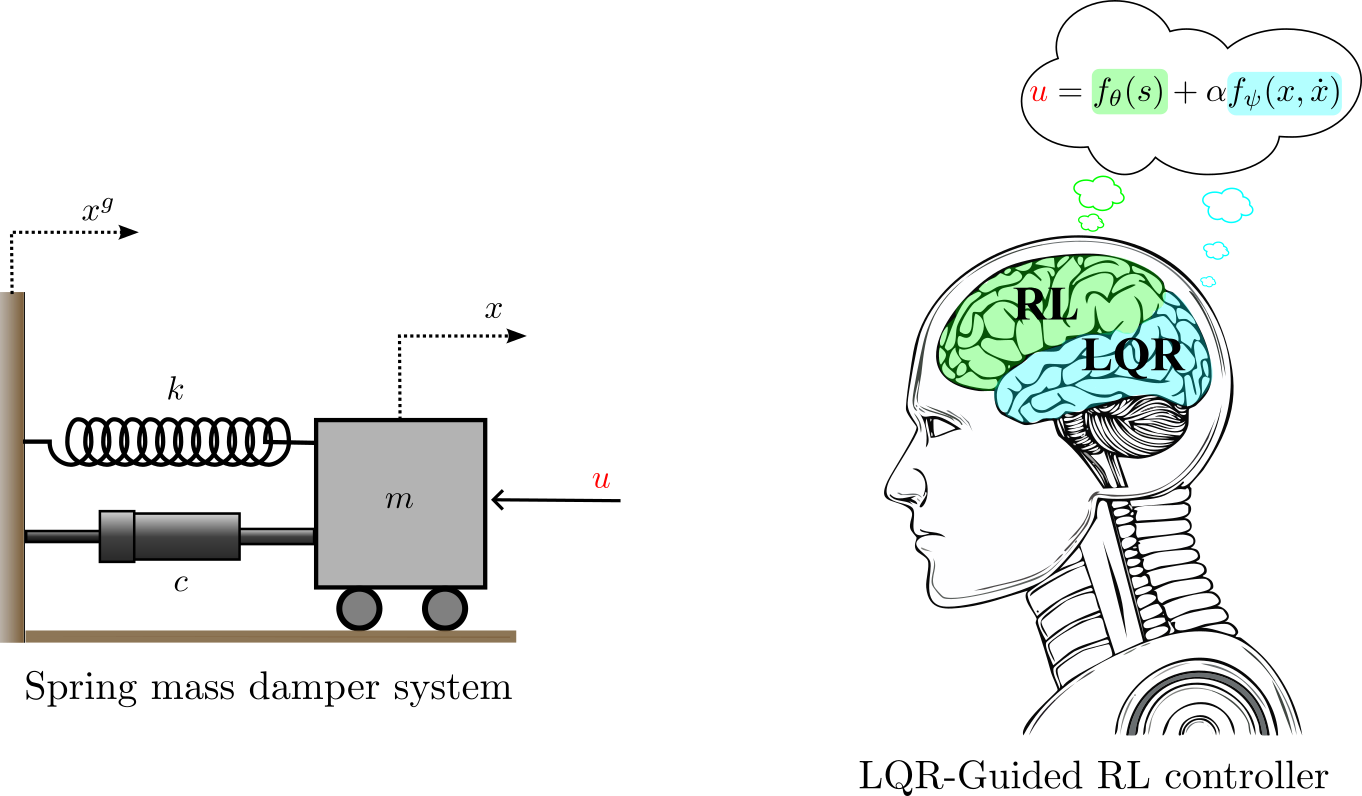}
    \caption{Overview of the LQR-Guided Reinforcement Learning framework: a classical spring-mass-damper system is controlled using a methodology that integrates Linear Quadratic Regulator (LQR) guidance within a reinforcement learning paradigm, leveraging both analytical control theory and data-driven learning.}
    \label{fig:LQR-Guided-RL-framework}
\end{figure}

Despite these advances, significant challenges remain. A key limitation is that most studies rely on training RL controllers in simulation environments that may not accurately capture the full dynamics of physical structures. For RL to be truly model-free and practically viable, controllers need to be trainable directly on real physical systems, which introduces challenges related to safety of the real physical system. This paper addresses these challenges by analyzing the feasibility of RL-based vibration control in realistic scenarios and proposing a practical solution of which a overview is shown in Figure \ref{fig:LQR-Guided-RL-framework}. 

The main contributions of this work are as follows: 
\begin{enumerate}
    \item We identify and highlight the key challenge of applying RL-based vibration control from a practical implementation perspective.  
    \item We propose a feasible solution to overcome these challenges and validate its effectiveness through a numerical case study. 
\end{enumerate}

\section{Background}
Classical approaches to control of dynamical systems typically rely on prior knowledge of the system dynamics. As a result, system identification often becomes an essential prerequisite to controller design. However, reinforcement learning (RL) offers an alternative framework, in which explicit identification of system dynamics is not required. In RL, the controller is modeled as an \textit{agent} that interacts directly with the dynamical system, referred to as the \textit{environment}. At each time step, the agent selects an action `$a$' based on a policy $f_\theta$, parameterized by $\theta$ and its state `$s$'. The environment then returns a state observation and a reward signal. These interactions are accumulated in a memory buffer, often termed the \textit{experience replay} $(D)$, which is subsequently sampled during training to improve the policy towards optimal performance. 
\begin{figure}
    \centering
    \includegraphics[width=0.75\linewidth]{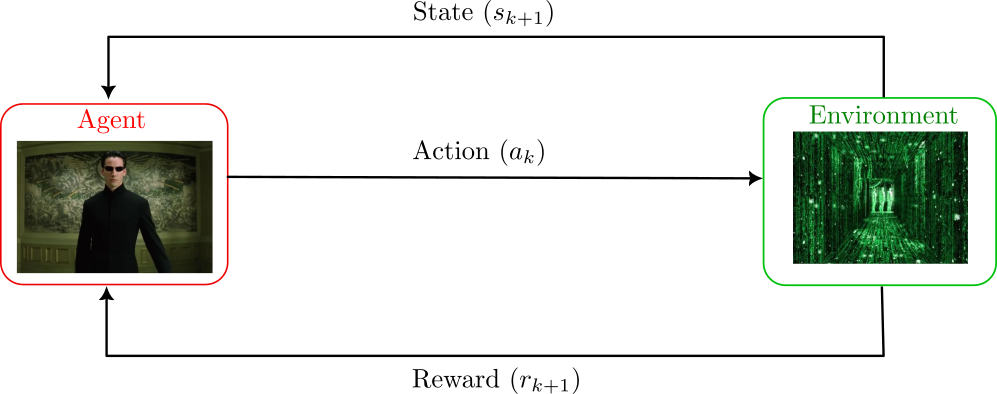}
    \caption{Overview of agent-environment interaction in reinforcement learning: The agent selects actions based on the observed state, receives rewards as feedback, and transitions to a new state through interaction with the environment}
    \label{fig:Agent_Env_interaction}
\end{figure}

In this work, a \textit{Lyapunov-based Actor-Critic (LAC)} method is employed. LAC is derived from the \textit{Soft Actor-Critic (SAC)} algorithm, which has been demonstrated to be effective for real-world applications, including benchmarks on quadrupedal robots as shown by Han et al.~\cite{han2020actor}. The distinguishing feature of LAC is the integration of a \textit{Lyapunov function} $L_\phi$, which enforces stability during the training process. In this method, the Lyapunov function serves as the \textit{critic}, while the RL agent operates as the \textit{actor}. The Lyapunov function is constrained to produce only positive real values and is defined as:  
\begin{equation}
L_\phi(s,a) = f_\phi(s,a)^\top f_\phi(s,a).
\end{equation}

The parameters of the Lyapunov function are optimized by minimizing the deviation between its estimated value and a target Lyapunov value, $L^{\text{target}}(s,a)$. The corresponding objective function is given by:
\begin{align}
    J(\phi) &= \mathbb{E}_D \left[ \tfrac{1}{2}\big(L_\phi(s,a) - L_{\phi'}^{\text{target}}(s,a)\big)^2 \right], \label{eq:phi_update}\\
    L_{\phi'}^{\text{target}}(s,a) &= -r + L_\phi(s',a'),
\end{align}

where $r$ is the reward obtained for executing action $a$ in state $s$. The target parameters $\phi'$ are updated using a Polyak averaging scheme:  

\begin{equation}
    \phi'_{k+1} \leftarrow \tau \phi_k + (1-\tau)\phi'_k.\label{eq:phi_dash_update}
\end{equation}

The policy parameters $\theta$ of the agent are updated by maximizing the following objective:  

\begin{align}
    J(\theta) = \mathbb{E}_D \Big[ \, \beta \big(\log(\pi_\theta(f_\theta(s)|s)) + \mathcal{H}_t\big) 
    + \lambda\big(L_\phi(s', f_\theta(s')) - L_\phi(s) + \alpha_3 c\big) \Big] \label{eq:theta_update}
\end{align}

where $\alpha_3$ is a tunable hyperparameter, while $\beta$ and $\lambda$ are adaptively optimized. The entropy term $\mathcal{H}_t$ encourages exploration, ensuring that the agent collects informative data across the state space. For a comprehensive exposition of the LAC method, readers are referred to Han et al.~\cite{han2020actor}. 

\newpage
\section{Practical concern of reinforcement learning implementation for vibration control}

Reinforcement learning (RL) has demonstrated superior performance compared to classical control strategies such as the Linear Quadratic Regulator (LQR), as reported in several recent studies \cite{eshkevari2023active,zhang2023novel,gheni2024intelligent,qiu2021reinforcement}. Despite its promising results, the practical deployment of RL-based controllers in vibration control systems presents significant challenges. A primary concern arises during the training phase of the RL policy. Since the agent learns by exploration, the generated control inputs are initially random. Consequently, the system experiences excessive oscillations and transient responses, which may pose a severe risk of structural damage if the training is conducted on a physical system.

\begin{figure}[!ht]
    \centering
    \includegraphics[width=1\linewidth]{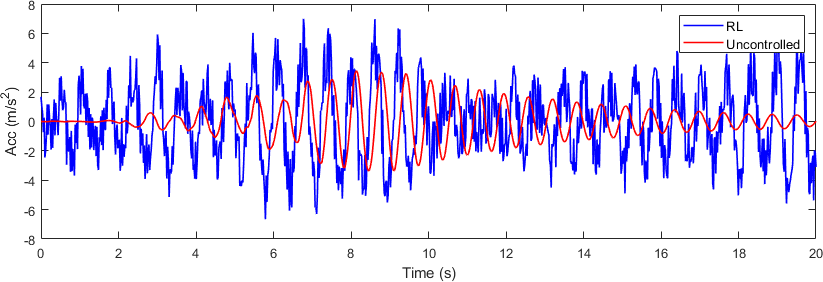}
    \caption{Comparison of acceleration response of a system during Reinforcement learning policy training and  the response of the uncontrolled system}
    \label{fig:RL_training}
\end{figure}

As illustrated in Figure \ref{fig:RL_training}, the acceleration response observed during the training phase is magnitudes higher as compare to uncontrolled system scenario. Such amplified vibrations are unacceptable for physical implementations, as they would induce significant wear or catastrophic system failure. The root cause is that during early training episodes, the RL controller exerts control forces with random characteristics, generated without domain-specific knowledge. 

This challenge creates a fundamental dilemma. On the one hand, training must be performed in a real environment to preserve the model-free property of RL, since reliance on an accurate simulator requires prior system identification. However, detailed system identification contradicts the motivation for adopting a model-free approach, and any inaccuracies in the identified model may compromise the learned policy’s transferability. To overcomes the observed/highlighted practical concern in case of vibration control using reinforcement learning a LQR-Guided RL approach is proposed and explained in detail in the next section.

\section{LQR-Guided RL framework}
In classical control approaches such as the Linear Quadratic Regulator (LQR), the control policy is derived from a assumed model for the dynamical system. Consequently, if the assumed model deviates significantly from the true system, the performance of the LQR controller can deteriorate. An interesting question arises: if the assumed model is substantially different from the actual system, will the LQR controller still mitigate system vibrations to some degree, or could it potentially destabilize or damage the system due to the erroneous control policy?

To investigate the robustness of the LQR method under model mismatch, consider a scenario involving a randomly assumed linear dynamical system and a true nonlinear dynamical system, both with a single degree of freedom, described as follows:
\begin{align*}
    &\text{Assumed linear system: } \quad \hspace{3em} m\Ddot{x} + c\Dot{x} + kx = u - m\Ddot{x}_{g} \\
    &\text{True nonlinear system: } \quad m_{\text{true}}\Ddot{x} + c_{\text{true}}\Dot{x} + k_{\text{true}}x + k'_{\text{true}}x^3= u - m_{\text{true}}\Ddot{x}^g
\end{align*}

where $x$ denotes the displacement response, $\Ddot{x}^g$ represents ground acceleration (generated via Kanai-Tajimi filter \cite{rofooei2001generation}), and $u$ is the control force input. The parameters value are: $m=1.6,c=-0.5,k=181,m_{\text{true}}=1,c_{\text{true}}=0.4,k_{\text{true}}=100$, and $k'_{\text{true}}=1$. Note that the assumed system parameters are selected arbitrarily without system identification, thus differing markedly from the true system. 

\begin{figure}[!ht]
    \centering
    \includegraphics[width=1\linewidth]{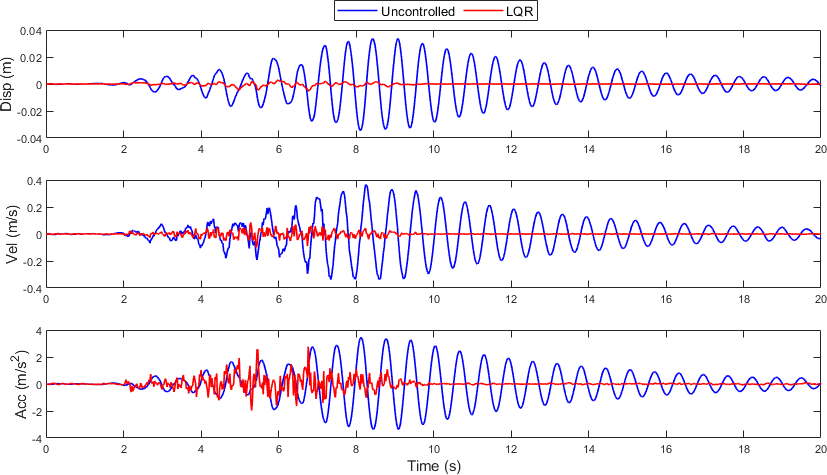}
    \caption{Comparison of displacement, velocity and acceleration response of the true nonlinear dynamical system for the case of LQR controller and uncontrolled scenario}
    \label{fig:LQR_vs_Uncontrolled}
\end{figure}

An LQR controller is designed based on the assumed linear system and subsequently tested on the true nonlinear system. The performance is compared with the uncontrolled case, as illustrated in Figure \ref{fig:LQR_vs_Uncontrolled}. Surprisingly, the LQR controller effectively reduces vibrations despite the model discrepancy. Although this outcome may seem counterintuitive initially, it can be rationalized by considering that both systems fundamentally behave as spring-mass-damper systems. Therefore, key dynamic characteristics, such as the spring’s restorative action opposing displacement, remain qualitatively same between assumed and true models.

From the results, it can be observed that an LQR controller designed using arbitrarily assigned system parameters, rather than an accurately identified model, effectively falls under the category of model-free control. Consequently, such an LQR policy can be utilized as a form of prior knowledge within the reinforcement learning (RL) framework. We define the LQR-guided RL policy as
\begin{equation*}
    u = f_{\theta}(s) + \alpha f_{\psi}(x,\Dot{x})
\end{equation*}
where $\theta$ and $\psi$ denote the parameters of the RL policy and the LQR policy, respectively. The hyperparameter $\alpha \in [0,1]$ regulates the weight of the LQR contribution in the combined LQR-guided RL policy. In the present study, $\alpha$ is fixed at 0.5. 

\begin{figure}[!ht]
    \centering
    \includegraphics[width=1\linewidth]{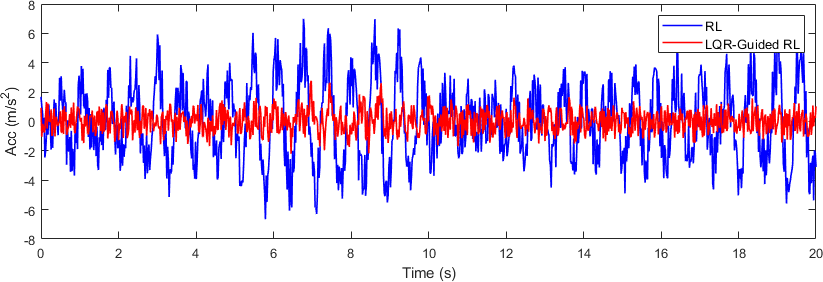}
    \caption{Comparison of acceleration response during training of RL controller and LQR-Guided RL controller}
    \label{fig:LQR_Guided_RL_training}
\end{figure}

\begin{figure}[!ht]
    \centering
    \includegraphics[width=1\linewidth]{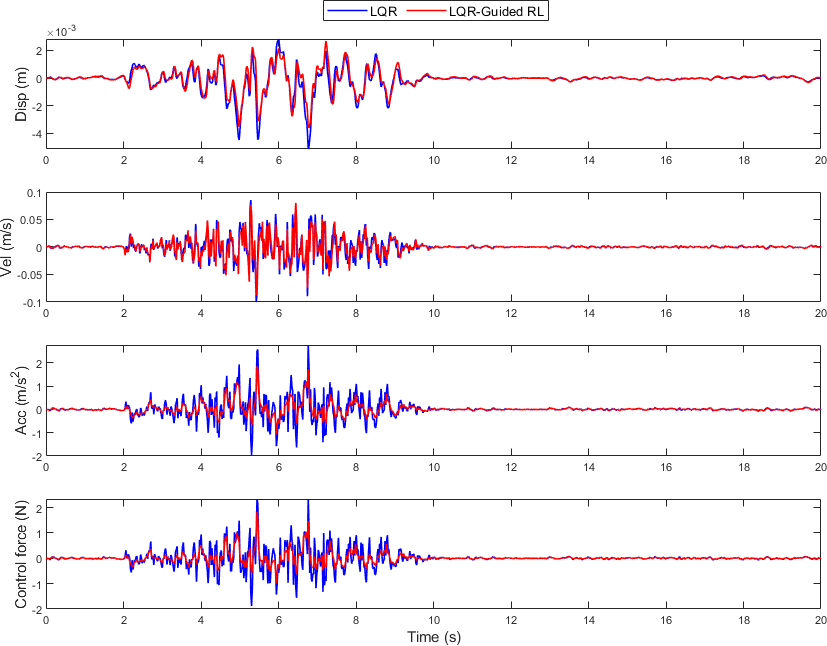}
    \caption{Comparison of displacement, velocity, acceleration response and control force during testing for the case of LQR-Guided RL and LQR}
    \label{fig:LQR_Guided_RL_testing}
\end{figure}

Here, the objective is to obtain a control policy that achieves both optimal performance during testing and minimal vibration during training. To assess the effectiveness of the proposed approach, the acceleration response of the dynamical system during the training phase is compared between the LQR-guided reinforcement learning (RL) policy and the non-guided RL policy. As shown in Figure \ref{fig:LQR_Guided_RL_training}, the incorporation of the LQR prior leads to a significantly reduced acceleration response relative to the non-guided RL policy. Furthermore, the LQR-guided RL policy and the standalone LQR policy are tested on the true nonlinear system. As illustrated in Figure \ref{fig:LQR_Guided_RL_testing}, the LQR-guided RL policy exhibits lower acceleration responses and requires reduced control effort compared to the LQR policy. These results indicate that LQR guidance improves the safety in the training process of the RL policy, and also, the trained RL policy achieves superior performance compared to the standalone LQR policy.

\begin{algorithm}
\caption{LQR-Guided RL controller training}
\label{alg: RL controller pseudo code}
\begin{algorithmic}[1] 
\STATE Design LQR policy $f_\psi$ based on randomly assumed linear dynamical system
\STATE Initialize parameter $\theta, \phi$ and $\phi' \leftarrow \phi$
\STATE Set reward weights $w_1,w_2,w_3$ and $T=20,\ dt=0.02, \alpha = 0.5$ 
\FOR{simulation = 1:100} 
    \STATE Initialize $x_1, \Dot{x}_1, s_1$ 
    \FOR{ $k=1:T/dt$}
    \STATE $\Tilde{u}_k = f_\theta(s_k)$
    \STATE $u_k = \Tilde{u}_k + \alpha f_\psi(x_k,\Dot{x}_k)$
    \STATE $x_{k+1},\Dot{x}_{k+1}, \Ddot{x}_{k+1} = f_{\text{env}}(x_k,\Dot{x}_k,u_k) $
    \STATE $u^*_{k+1} = f_\psi(x_{k+1},\Dot{x}_{k+1})$
    \STATE $r_{k+1} = -\big[w_1 ||x_{k+1}|| + w_2 ||\Ddot{x}_{k+1}|| + w_3 ||u_k|| \big]$
    \STATE $s_{k+1} = \{\Ddot{x}_{k-l+2:k+1},\Ddot{x}^g_{k-l+2:k+1},\Tilde{u}_{k-l+1:k},u^*_{k-l+2:k+1}\}$
    \STATE store $s_k,u_k,r_{k+1},s_{k+1}$ in experience replay $D$
    \STATE update $\theta,\phi,\phi'$ using Eq. \ref{eq:theta_update}, \ref{eq:phi_update}, and \ref{eq:phi_dash_update}
    \ENDFOR

\ENDFOR
\RETURN control policy parameter $\theta,\psi$
\end{algorithmic}
\end{algorithm}

\newpage
\section{Summary and future work}
The LQR-Guided RL framework presented in this work directly addresses the practical safety challenges of training RL controllers on physical structures by leveraging prior knowledge from LQR policies, even when based on arbitrarily assumed models. The results demonstrate that the proposed approach reduces training-related risks compared to naive model-free RL, with the final RL-trained controller outperforming a standalone LQR controller. Notably, this method achieves these advantages while remaining model-free, thus eliminating the need of system identification and offering a safe control framework for structural vibration mitigation.

For future work, the viable directions are as follows:
\begin{itemize}
    \item \textbf{Extension to MultiDOF Systems:} Applying and validating the LQR-guided RL approach for multi-degree-of-freedom structural systems to enable its use in more complex, realistic engineering scenarios.
    \item \textbf{Experimental Validation:} Conducting hardware implementation and real-time experiments to confirm safety and performance advantages observed in simulations.
    \item \textbf{Reduction of Training Duration:} Developing methods to reduce RL training time, thereby enhancing the practical viability of this safe learning approach for large-scale systems.
\end{itemize}
\begin{table}[ht]
\centering

\begin{tabular}{ll}
\hline
\textbf{Hyperparameter}                            & \textbf{Values / Range}           \\ \hline
Learning rate of actor                                       & $10^{-4}$                    \\
Learning rate of critic                                       & $3\times10^{-4}$                    \\
Actor network architecture                 & 4 layers, 256-256-256-1 neurons                 \\
Critic network architecture                 & 4 layers, 256-256-256-1 neurons                 \\
Activation function                     & Leaky ReLU                      \\
Batch size                             & 256                               \\
Discount factor ($\gamma$)                           & 0.998                               \\ 
Length of response history ($l$) & 4\\
    Rewards weights $[w_1,w_2,w_3]$       & $[1,10^{-2},10^{-3}]$          \\
Optimizer                      & Adam                        \\
Number of simulations (epochs)                & 100                                 \\
Function $f_\theta$ output range & [-1, 1]\\ \hline
\end{tabular}
\vspace{1em}
\caption{Hyperparameter of the LQR-Guided RL controller}
\label{tab:rl_hyperparameters}
\end{table}

\section*{Code Availability}
The code that supports the findings of this study is openly available at the following GitHub repository: \href{https://github.com/rohan-v-thorat/LQR-Guided-RL}{https://github.com/rohan-v-thorat/LQR-Guided-RL}. The repository includes all scripts, functions, and instructions necessary to reproduce the results presented in this paper.

%
%

\end{document}